\documentclass[lettersize,journal]{IEEEtran}
\usepackage{amsmath,amsfonts}
\usepackage{algorithmic}
\usepackage{algorithm}
\usepackage{array}
\usepackage[caption=false,font=normalsize,labelfont=sf,textfont=sf]{subfig}
\usepackage{textcomp}
\usepackage{stfloats}
\usepackage{url}
\usepackage{verbatim}
\usepackage{graphicx}
\usepackage{cite,xcolor}
\usepackage{multirow}
\usepackage{booktabs}
\usepackage{xcolor}
\usepackage{subfig}

\begin{document}

\title{Mixture-of-Noises Enhanced Forgery-Aware Predictor for Multi-Face Manipulation Detection and Localization}

\author{Changtao Miao, Qi Chu, Tao Gong, Zhentao Tan, Zhenchao Jin, Wanyi Zhuang, Man Luo, Honggang Hu, and Nenghai Yu
\thanks{
Email: miaoct@mail.ustc.edu.cn.
}
% \thanks{
% This work was supported by the National Natural Science Foundation of China (No. 62121002) and the Fundamental Research Funds for the Central Universities.
% %Correspondence to: Qi Chu.

% Changtao Miao, Qi Chu, Tao Gong, Wanyi Zhuang, Honggang Hu, and Nenghai Yu are with the School of Cyber Science and Technology, University of Science and Technology of China, Hefei 230026, China, and also with the Anhui Province Key Laboratory of Digital Security, Hefei 230026, China (email: \{miaoct, wy970824\}@mail.ustc.edu.cn; \{qchu, tgong, hghu2005, ynh\}@ustc.edu.cn).

% Zhentao Tan is with Alibaba Cloud, Hangzhou 311121, China (email: tzt@mail.ustc.edu.cn).

% Zhenchao Jin is with the Department of Statistics and Actuarial Science, The University of Hong Kong, Hong Kong 999077, China (email: blwx96@connect.hku.hk).

% Man Luo is with Ant Group, Hangzhou 310023, China (email: sankuai.luoman@antgroup.com).
% }
}
\markboth{Journal of \LaTeX\ Class Files,~Vol.~14, No.~8, August~2021}%
{Shell \MakeLowercase{\textit{et al.}}: A Sample Article Using IEEEtran.cls for IEEE Journals}


\maketitle

\begin{abstract}
With the advancement of face manipulation technology, forgery images in multi-face scenarios are gradually becoming a more complex and realistic challenge. 
Despite this, detection and localization methods for such multi-face manipulations remain underdeveloped.
Traditional manipulation localization methods either indirectly derive detection results from localization masks, resulting in limited detection performance, or employ a naive two-branch structure to simultaneously obtain detection and localization results, which cannot effectively benefit the localization capability due to limited interaction between two tasks.
This paper proposes a new framework, namely \textit{MoNFAP}, specifically tailored for multi-face manipulation detection and localization.
The \textit{MoNFAP} primarily introduces two novel modules: the Forgery-aware Unified Predictor (FUP) Module and the Mixture-of-Noises Module (MNM). The FUP integrates detection and localization tasks using a token learning strategy and multiple forgery-aware transformers, which facilitates the use of classification information to enhance localization capability. 
Besides, motivated by the crucial role of noise information in forgery detection, the MNM leverages multiple noise extractors based on the concept of the mixture of experts to enhance the general RGB features, further boosting the performance of our framework.
Finally, we establish a comprehensive benchmark for multi-face detection and localization and the proposed \textit{MoNFAP} achieves significant performance. The codes will be made available.
\end{abstract}

\begin{IEEEkeywords}
Multi-face Manipulation, Face Manipulation Localization, Mixture of Experts, Masked Attention.
\end{IEEEkeywords}

\section{Introduction}
\IEEEPARstart{F}{ace} manipulation technologies \cite{faceswap2019,nirkin2019fsgan,perov2020deepfacelab,shen2020interpreting,pidhorskyi2020adversarial} rapidly evolve, achieving increasingly realistic results. 
Recently, driven by real-world demand, the focus has shifted from single-face forgery to multi-face scenarios \cite{le2021openforensics,zhou2021face,haiwei2022exploring}, which gives rise to serious malicious abuses, such as misinformation and fraud. 
Multi-face manipulation images possess the remarkable ability to manipulate and alter the semantic identity or expression attributes of one or multiple faces, increasing the difficulty of detection and introducing new challenges in localizing partial tampering regions. 
Consequently, developing effective methods for multi-face manipulation detection and localization is crucial.

%--------------Figure motivation---------------
\begin{figure}[!t]
\centering
\includegraphics[width=0.42\textwidth]{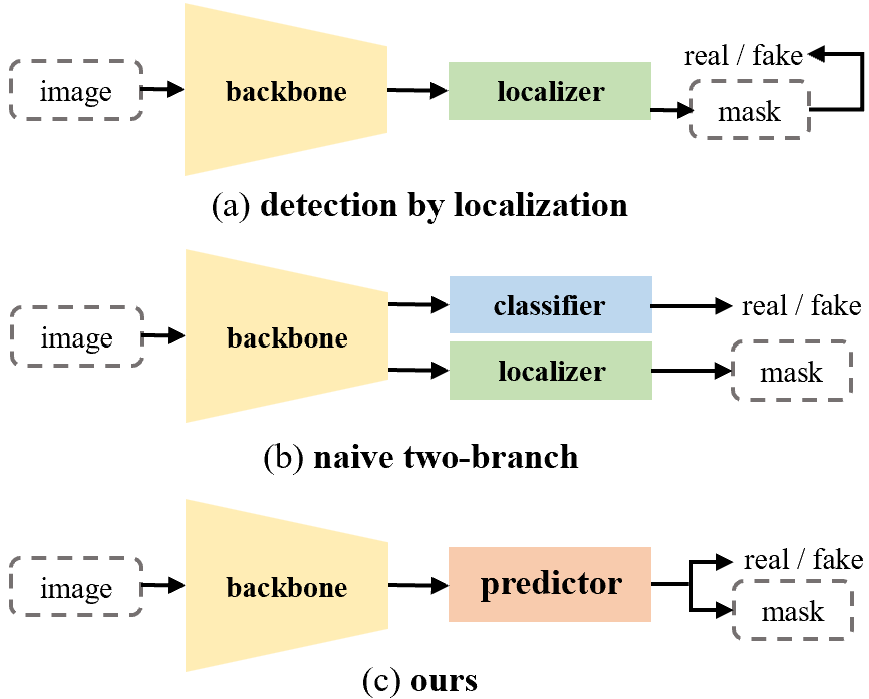}
\caption{Three different paradigms: (a) detection-by-localization which indirectly obtains image-level detection result from the pixel-level mask, (b) two-branch architecture employs separate classification branch and localization branch with shared backbone, and (c) our unified framework which integrates the detection and localization processing into a single predictor.
}
\label{fig:motivation_1}
\end{figure}
%--------------Figure motivation---------------

In recent developments, several methods \cite{li2020sharp,zhou2021face,lin2023exploiting,coccomini2022mintime} have made notable efforts to tackle the challenge of detecting multi-face manipulation images. 
However, these approaches predominantly emphasize image-level classification, which cannot precisely localize tampered face regions at the pixel level. 
To address the multi-face forgery localization, \cite{le2021openforensics} introduces a pioneering multi-face dataset called OpenForensics and leverages a pipeline based on instance segmentation \cite{he2017mask} to tackle these challenges. Unfortunately, this dataset solely consists of forged data and lacks corresponding genuine images. This has resulted in recent research work\cite{le2021openforensics,zhang2024comics} based on this dataset being limited to localizing tampered face regions within the manipulated images while unsuitable for performing image-level fake/real detection.
In real-world scenarios, both image-level detection and pixel-level localization play crucial roles in analyzing multi-face forgery data. However, the simultaneous solution of both tasks remains under-explored.
%a challenging problem.

\begin{figure}[!t]
\centering
\includegraphics[width=0.48\textwidth]{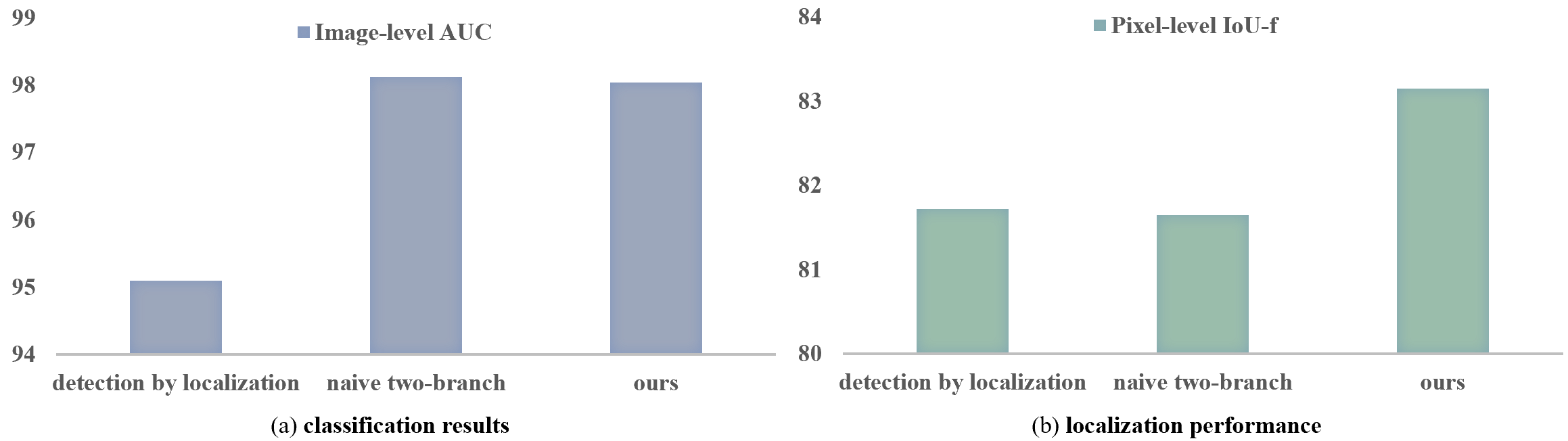}
\caption{(a) The detection-by-localization method shows limited image-level detection performance. In contrast, two-branch and our methods can release the potential of the model’s detection capabilities.
(b) The two-branch approach cannot effectively improve the localization performance compared to the detection-by-localization counterpart, while our method can facilitate the use of classification
information to enhance localization capability. More experimental results and analyses are shown in Tab. \ref{tab:ab_modes} of Sec. \ref{sec_Fab_1}.
}
\label{fig:motivation_2}
\end{figure}

Conventional image forensics methods \cite{wu2019mantra,mayer2019forensic,niu2021image,zhuang2021image} employ detection-by-localization paradigm to obtain image-level detection result from pixel-level mask indirectly (see Fig. \ref{fig:motivation_1}(a)). The detection result heavily relies on the quality of localization, resulting in limited detection performance (see Fig. \ref{fig:motivation_2}(a)). Some recent image forensics methods \cite{liu2022pscc,guo2023hierarchical,islam2020doa,wang2022objectformer,guillaro2023trufor} utilize two-branch architecture to obtain image-level detection and pixel-level localization results simultaneously (see Fig. \ref{fig:motivation_1}(b)), which releases the potential of the model's detection capabilities. However, the localization capability can hardly benefit from the design of a separate classification branch and localization branch with a shared backbone (see Fig. \ref{fig:motivation_2}(b)), due to limited interaction between the two tasks.

%--------------Figure motivation---------------
%--------------Figure motivation---------------

In this paper, we propose a Forgery Perception Unified Predictor (FUP) to predict pixel-level localization and image-level classification results jointly (see Fig. \ref{fig:motivation_1}(c)), which can effectively utilize the classification information to enhance the localization capability. 
Specifically, the FUP introduces a token learning strategy alongside multiple Forgery-aware Transformer (FAT) modules, establishing a robust connection between detection and localization tasks. 
Within the FAT module, we incorporate two learnable tokens that represent real and fake categories. These tokens, along with image features, are updated bidirectionally through token self-attention and token-image cross-attention (i.e., tokens to image feature, and vice-versa), effectively encoding global contextual information.
Then, by employing direct image-level classification through supervised learning, the real-fake tokens enhance their ability to capture and express critical information in forged images, particularly in relation to manipulated regions. 
This capability significantly improves mask prediction accuracy within the FAT module.
Additionally, we implement forgery-aware masked attention within the FAT module, which significantly constrains cross-attention to the forgery and non-forgery regions of the predicted mask for fake and real query tokens, respectively. 
This innovative approach allows the global real-fake tokens to concentrate on microscopic forgery cues, enhancing the model's sensitivity to subtle manipulations.
Meanwhile, considering the prevalence of partial and subtle forged face regions, the FUP employs a multi-scale strategy to extract features across a feature pyramid structure. 
Finally, FUP can simultaneously produce the final set of detection and localization results by reasoning about the relationships between output tokens and image features. 
In contrast to the prior work, our FUP streamlines the detection and localization tasks pipeline and leverages the complementary information from both tasks, particularly enhancing localization performance.

Besides, considering that the previous various noise extractors \cite{li2019localization,fridrich2012rich,bayar2018constrained,yu2020searching,yang2021mtd} which
have shown impressive results in the field of image forensics and deepfake detection, we introduce the Mixture of Noise Extractors (MoNE) module, inspired by the mixture-of-experts (MoE) philosophy \cite{shazeer2017outrageously}, to harness the benefits of diverse noise types for augmenting the forgery cue patterns of the RGB features.  
The MoNE module employs various noise extractors to extract diverse and comprehensive forgery cues to the plain image features.
Unlike independent noise extractors, our MoNE module attempts to exploit a combination of noise information with different properties during training. 
To naturally adapt to the architecture of the FUP, we propose the Mixture-of-Noises Module (MNM) processes features across various resolutions by leveraging multiple MoNE modules.
Subsequently, each resolution of the multi-scale noise patterns is fed into a corresponding FAT layer of the FUP, which aids the model in localizing small forgery regions.

Meanwhile, since existing multi-face related datasets either lack corresponding real data \cite{le2021openforensics} or are not suitable for multi-face detection and localization tasks \cite{zhou2021face,haiwei2022exploring},  the multi-face manipulation detection and localization community currently lacks a well-developed multi-face forgery benchmarks. To bridge this gap, we collect multi-face data from existing datasets to curate multi-face manipulation detection and localization benchmark datasets comprising diverse real-world scenarios, such as movies, plays, news broadcasts, variety shows, and interviews.
Based on the curated dataset, we construct comprehensive benchmarks for evaluating the generalization and robustness of multi-face manipulation detection and localization methods.

Overall, in this work, we make the following key contributions:
\begin{itemize}
  \item We propose a unified framework, namely \textbf{MoNFAP}, for multi-face manipulation detection and localization tasks, which primarily comprises two novel modules: the Forgery-aware Unified Predictor (FUP) Module and the Mixture-of-Noises Module (MNM). 
  \item The proposed FUP integrates detection and localization tasks using a token learning strategy and multiple forgery-aware transformers, which facilitates the use of classification information to enhance localization capability. 
  \item The proposed MNM, inspired by the concept of the MoE, leverages multiple noise expert extractors to extract more general and robust noise traces, thus enhancing the plain image features of the FUP.
  \item We construct comprehensive multi-face manipulation detection and localization benchmarks and the proposed method achieves state-of-the-art performance.
\end{itemize}

%--------------Figure framework---------------
\begin{figure*}[!t]
\centering
\includegraphics[width=0.99\textwidth]{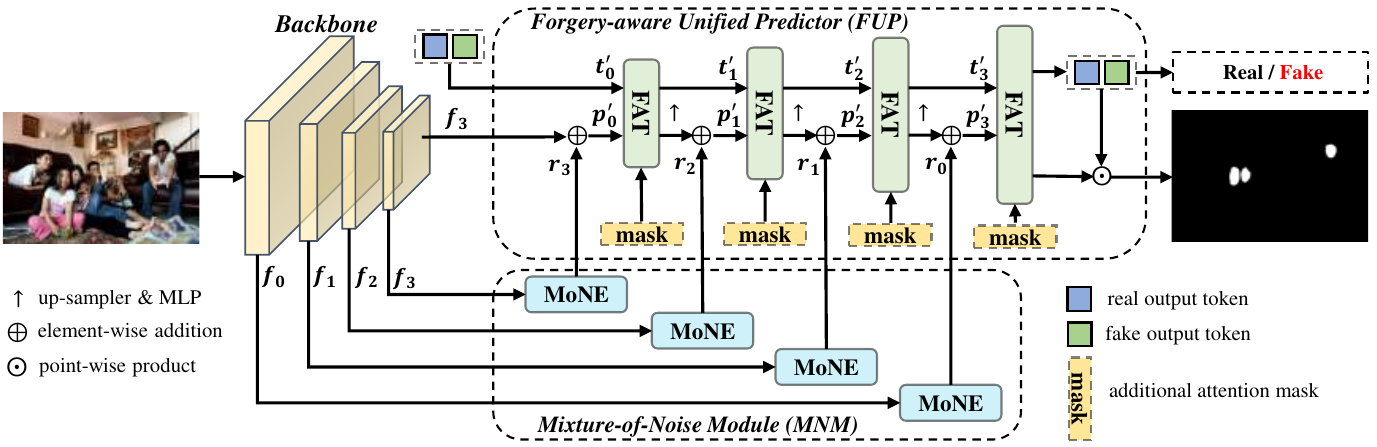}
\caption{Detailed architecture of the proposed MoNFAP. Firstly, we employ the MoNP module, which consists of four Mixture of Noise Extractors (MoNE) modules. These MoNE modules process multi-scale features obtained from the backbone network. The MNM outputs noise patterns that enhance the general features within the FUP. Lastly, the FUP module utilizes the output tokens and the Forgery-aware Transformer (FAT) to jointly predict classification and localization results. To maintain clarity, we omit the two outputs of FAT and the generation of the auxiliary layer for the attention mask.
}
\label{fig:framework}
\end{figure*}
%--------------Figure framework---------------

\section{Related Work}

\subsection{Face Manipulation Detection and Localization}
Early approaches in the field utilized hand-crafted biological cues \cite{li2018ictu} derived from CNNs to distinguish between real and fake faces. 
However, contemporary methods predominantly adopt data-driven approaches \cite{fernando2020detection,yang2021mtd,wang2022forgerynir,yang2020preventing,song2022face,miao2021towards,miao2021learning,zhuang2022towards,yang2023avoid,li2023forensic,xie2023domain}, training deep networks directly on real and fake images. 
Furthermore, certain techniques exploit generalized artifacts in the frequency domain \cite{miao2022hierarchical,tan2022transformer,wang2022lisiam,miao2023f,guo2023constructing,liu2023adaptive}, inconsistencies \cite{zhuang2022uia}, or spatial patterns \cite{yin2023dynamic,han2023fcd,liu2023fedforgery,zhao2023istvt,yang2023masked,luo2023beyond} mechanisms to enhance the overall performance of face forgery detection.
Recent studies \cite{li2020sharp,zhou2021face} address multi-face forgery detection using multiple-instance learning and video-level labels. 
FITER \cite{lin2023exploiting} leverages facial relationships within multiple faces to enhance multi-face forgery detection. However, these methods lack pixel-level localization of tampered face regions. 

Previous FFD \cite{dang2020detection} uses the network's attention map for tampered region detection but lacks global context. Other methods \cite{jia2021inconsistency,kong2022detect} incorporate a segmentation branch for localizing manipulated regions. Existing localization method \cite{huang2022fakelocator} for full-synthesized fake images are not suitable for face manipulation data. Recent approaches \cite{le2021openforensics,zhang2024comics} leverage instance segmentation pipelines to localize tampered and authentic face regions for multi-face manipulation images but do not address image-level real/fake classification. 
MSCCNet \cite{miao2023multi} learns semantic-agnostic features through multi-spectral information.
In this paper, we propose MoNFAP to efficiently detect and localize multi-face forgery images by simultaneously processing multiple faces from an input image.

\subsection{Manipulation Noise Artifacts}
Low-level artifacts in image editing and tampering can be highlighted by noise extraction modules. These modules transform the input image from RGB space to a semantic-agnostic noise space by suppressing semantic content. HFConv \cite{li2019localization} introduces trainable high-pass filters for image forensics, and frequency-domain information has been effective in face manipulation detection \cite{wang2022objectformer,wang2022m2tr}. SRMConv \cite{fridrich2012rich,wu2019mantra} learns edge and boundary features without relying on pre-defined manipulation artifacts, making it suitable for noise-sensitive analysis. BayarConv \cite{bayar2018constrained} enables direct learning of manipulation traces during training to extract forgery noise patterns, employed in numerous approaches \cite{wu2019mantra,chen2021image}. CDConv \cite{yu2020searching} utilizes the central difference operator to capture forgery cues and representations in face manipulation detection \cite{yang2021mtd}. However, existing methods typically use only one or two noise extractors, failing to exploit their potential advantages fully and resulting in sub-optimal performance.
Moreover, noise extractors are often used as data augmentation techniques or incorporated within single-level features, overlooking multi-level features and valuable information for enhanced detection and localization accuracy. In this paper, we propose the Mixture of Noise Extractors (MoNE) modules integrated at multi-level to improve the generalization capability of semantic-agnostic tampering trace features.

\subsection{Image Manipulation Detection and Localization}
In image forensics, integrating image-level detection and pixel-level localization is crucial for real-world applications. Most existing methods focus on localization only, ignoring image-level classification \cite{li2019localization,zhou2023pre,zhuo2022self,wang2023shrinking}. Previous approaches compute the classification score by extracting a global decision statistic from the localization mask, prioritizing localization over detection \cite{wu2019mantra,niu2021image,zhuang2021image,guillaro2023trufor}. Some recent methods address detection explicitly by incorporating authentic data and using image-level classification losses, but they may hinder the learning of high-level semantic information and result in subpar classification performance \cite{chen2021image}. Other approaches \cite{liu2022pscc,guo2023hierarchical,islam2020doa,wang2022objectformer,guillaro2023trufor} introduce additional classification branches but do not fully explore the feature-level interaction between classification and localization tasks. In this paper, we propose the Forgery-aware Unified Predictor (FUP), which optimizes the detection and localization pipeline and leverages classification information to enhance localizer performance.

\section{Method}
In this section, we first provide a comprehensive overview of the MoNFAP framework (in Sec.\ref{sec_overview}).
Then we introduce the Forgery-aware Unified Predictor (in Sec.\ref{sec_FUP})  and a detailed presentation of the Mixture-of-Noises Module (in Sec.\ref{sec_MNP}), respectively. 
Finally, we describe the Loss Function (in Sec.\ref{sec_loss}) employed in our framework to optimize the model's performance.

\subsection{Overview} \label{sec_overview}
% \jzc{
As depicted in Figure \ref{fig:framework}, the proposed MoNFAP framework comprises three primary components: a backbone network, a Mixture-of-Noises Module, and a Forgery-aware Unified Predictor. 
Given an input image $I \in \mathbb{R}^{3 \times H \times W}$, the backbone network is first utilized to extract multi-scale features $\mathbf{F}=\{f_0 \in \mathbb{R}^{C \times \frac{H}{4} \times \frac{W}{4}},  f_1 \in \mathbb{R}^{2C \times \frac{H}{8} \times \frac{W}{8}}, f_2 \in \mathbb{R}^{4C \times \frac{H}{16} \times \frac{W}{16}}, f_3 \in \mathbb{R}^{8C \times \frac{H}{32} \times \frac{W}{32}}\}$, where $H \times W$ denotes the image resolution and $C$ signifies the feature channels. 
Following this, the Mixture-of-Noises Module is structured to utilize four distinct Mixture of Noise Extractors (MoNE) to process $\mathbf{F}$ for the extraction of associated noise features denoted as $\mathbf{R} = \{r_0, r_1, r_2, r_3\}$. Notably, $\mathbf{R}$ shares the same dimensions as $\mathbf{F}$ and is intended to function as forgery noise cues for the subsequent Forgery-aware Unified Predictor.
Specifically, our predictor is structured into four stages, with each stage being composed of a Forgery-aware Transformer (FAT) layer. 
For each stage $i \in [0, 3]$, the following equation governs the process,
\begin{equation}
    \mathbf{t}_{i+1}, \textbf{p}_{i+1} = \textbf{FAT}_{i} (\mathbf{t}'_{i}, \textbf{p}'_{i}, mask),
\end{equation}
where $\textbf{t}_i$ comprises two learnable tokens, namely a real token and a fake token, $\textbf{t}'_i = \textbf{MLP}_i (\textbf{t}_i)$, $\textbf{p}'_i = \textbf{UP}_i (\textbf{p}_i) \oplus r_{3-i}, s.t. i \ge 1$ and $mask$ is a coarse predicted segmentation mask originating from an auxiliary layer. The symbol $\oplus$ represents the element-wise addition operation. The $\textbf{MLP}_i$ denotes a fully connected layer utilized to synchronize the feature channels of the learnable tokens and $\textbf{UP}_i$ denotes an up-sampler to synchronize the shape of the image features at stage $i$.
At the commencement of the process, we perform a random initialization for $\textbf{t}'_0 \in \mathbb{R}^{2 \times 8C}$, concurrently set $\textbf{p}'_0 = f_3 \oplus r_3$. 
The architecture of $\textbf{FAT}_i$ primarily comprises self-attention and cross-attention layers, facilitating the interaction and updating of $\textbf{t}'_{i}$ and $\textbf{p}'_i$.
Finally, $\textbf{t}_{4}$ and $\textbf{p}_4$ are utilized to derive both the image-level classification result $Y \in \mathbb{R}^{2} $ and a pixel-level localization mask $M \in \mathbb{R}^{2\times \frac{H}{4} \times \frac{W}{4}}$ in the following manner, 
\begin{equation}
Y = \textbf{MLP}(avg(\textbf{t}_{4})),
\end{equation}
\begin{equation}
M = \textbf{t}_{4} \odot \textbf{p}_4,
\end{equation}
where $avg$ indicates the average-pooling operation and $\odot$ means the spatially point-wise product.

\subsection{Forgery-aware Unified Predictor}
\label{sec_FUP}
In this paper, we propose a direct set prediction approach, namely Forgery-aware Unified Predictor (FUP), which comprises two learnable output tokens, four Forgery-aware Transformer (FAT) layers, and three up-sample layers, as illustrated in Fig. \ref{fig:framework}. 
To effectively handle small forgery regions, we employ a multi-scale strategy to feed successive noise cue features (see Sec.\ref{sec_MNP}) from different stages into successive FAT layers in a round-robin fashion, allowing the model to capture fine-grained forgery details. 
The successive FAT layers efficiently map image features, noise prompt features, and learnable output tokens to generate local forgery region masks and real-fake classification results. 
The proposed FUP thereby evolves into a feature pyramid structure, enabling the processing of noise cues and image features at different resolutions. 
In the following sections, we provide a detailed explanation of these improvements.

%--------------Figure FAT MoNE---------------
\begin{figure*}[!t]
\centering
\includegraphics[width=0.96\textwidth]{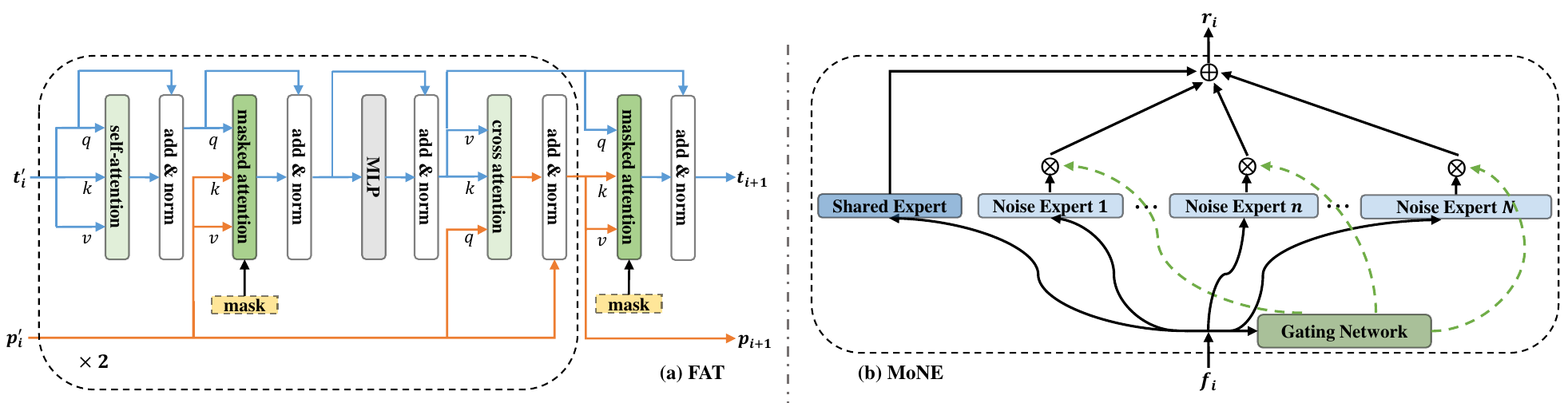}
\caption{Detailed architecture of the Forgery-aware Transformer (FAT) and Mixture of Noise Extractor (MoNE) modules.
(a) The blue and orange lines represent the output tokens and image features computation flow, respectively. The $\times2$ indicates that the computation is repeated twice.
(b) In the MoNE module, the $\oplus$ denotes element-wise addition, while the $\otimes$ represents element-wise multiplication. The dashed line indicates the output of the adaptive weight computed by the gating network.
% controlling the number of samples through the experts.
}
\label{fig:fat}
\end{figure*}
%--------------Figure FAT MoNE---------------

\subsubsection{Forgery-Aware Transformer (FAT)}
Inspired by the achievements of transformer-based architectures \cite{cheng2021per,cheng2022masked}, we find that real-fake categories regions in a multi-face manipulation image can be represented as object queries (i.e., real-fake output tokens). 
Thus, we introduce two learnable output tokens, namely the real token and the fake token, mathematically represented as $\textbf{t}'_i, i \in [0,3]$.
A transformer network can process the learnable tokens and image features to predict localization and classification results.
To this end, we propose the Forgery-aware Transformer (FAT) module that employs vanilla self-attention, masked cross-attention, and vanilla cross-attention in two directions (token-to-image embedding and vice versa) to process output tokens ($\textbf{t}'_i$) and enhanced image features ($\textbf{p}'_i$), as illustrated in Fig. \ref{fig:fat}(a).
After the above two operations, we again use masked cross-attention to make the real-fake tokens focus more on image features.

The key component of our FAT is a masked attention mechanism. Masked attention excels at extracting localized features by confining cross-attention solely within the manipulated region of the predicted mask for each object query, eschewing the conventional practice of attending to the entire feature map.
Specifically, this mechanism ensures that attention is only applied within the foreground region of the predicted mask for each query. Mathematically, this can be expressed as:
% --------Eq:masked-attn-----------
\begin{equation} \label{eq:masked-attn}
    q = Softmax(\textbf{Mask} + q{k^T})v,
\end{equation}
\begin{equation}
\textbf{Mask}(m,n) = \begin{cases}
            0 & \text{if $mask(m,n) = 1$ } \\
            -\infty & \text{otherwise}
        \end{cases},
\end{equation}
% --------Eq:masked-attn-----------
in which, $q$ refers to real-fake output tokens and $k,v$ are the image features. 
The coordinates of the feature location are denoted as $(m,n)$. The $mask$ binarizes with a threshold of 0.5, is obtained from the resized mask prediction of the auxiliary localizer.

The global context plays a crucial role in image segmentation tasks \cite{cheng2021per,cheng2022masked}. Still, it often contains an abundance of semantic objective features, which can harm semantic-agnostic forged regions \cite{chen2021image,sun2023safl}.
Therefore, we introduce a masked cross-attention mechanism to enhance the focus on local forged regions within the transformer module and mitigate the influence of the background global context. 

\subsubsection{Multi-scale Strategy (MSS)}
In multi-face manipulation images, the region of the forged face usually accounts for a smaller percentage compared to the whole image, making the modeling of local and subtle forged features very challenging. 
To better handle small forged face regions, we introduce a multi-scale strategy to enhance RGB image features using noise cues ranging from low to high resolution. 
Specifically, we utilize the multi-scale noise cues produced by the MNM (see Sec. \ref{sec_MNP}) with resolutions of $1/32$, $1/16$, $1/8$, and $1/4$ of the original RGB image as inputs to the FUP. 
Different scales of noise cues are added element-wise with the corresponding RGB image features in the FUP to enhance the forgery cues, as shown in Fig. \ref{fig:framework}. 
At each scale, the enhanced RGB image features and output tokens are updated by the FAT module for input to the next stage. 
This multi-scale procedure ensures effective information exchange between the output tokens, noise cues, and RGB image features. 
The FUP globally reasons about all objects together using pair-wise relations between them, while leading to enhanced representation and discriminative power in forgery detection and localization tasks. 
After running successive multi-scale FAT layers, a linear MLP maps the final real-fake output tokens to image-level classification, while a spatially point-wise product between the final RGB image features and real-fake output tokens produces the pixel-level mask.

\subsection{Mixture-of-Noises Module}\label{sec_MNP}
Inspired by the remarkable capabilities demonstrated by Mixture-of-Experts approaches \cite{jacobs1991adaptive,shazeer2017outrageously,dai2024deepseekmoe} in terms of computational efficiency and representation learning, we introduce a specially crafted Mixture of Noise Extractors (MoNE) module. This module enables the extraction of diverse forgery cues by leveraging different noise extractors comprehensively.
To capture noise patterns across various levels of features, we seamlessly integrate the MoNE module at multiple stages within the MoNFAP framework (as shown in Fig. \ref{fig:framework}). 
This integration leads to the creation of the Mixture-of-Noises Module, enhancing local forgery representation within the general RGB features of the FUP, while effectively suppressing the influence of semantic object content information.
In the following subsections, we delve into the detailed architecture and functionality of these modules.

\subsubsection{Preliminaries of the Mixture-of-Experts (MoE)}
The widely adopted architecture of the MoE \cite{shazeer2017outrageously} is commonly utilized in language modeling and machine translation tasks. It comprises a collection of $N$ expert networks denoted as $\{E_1, \cdots, E_n, \cdots, E_N\}$, along with a softmax gating network $G$. When provided with an input $x$, the base MoE layer produces an output $y$ that can be expressed as follows:
% --------Eq:moe-----------
\begin{equation}\label{eq:moe}
y = \sum_{n=1}^{N}G(x)_nE_n(x),
\end{equation}
% --------Eq:moe-----------
% --------Eq:gate-----------
\begin{equation}\label{eq:gate}
G(x) = Softmax(TopK(H(x), k)),
\end{equation}
% --------Eq:gate-----------
% --------Eq:noise-----------
\begin{equation}\label{eq:noise}
H(x)_n = (x \cdot W_g)_n + SN() \cdot Softplus((x \cdot W_{noise})_n),
\end{equation}
% --------Eq:noise-----------
% --------Eq:keeptopk-----------
\begin{equation}\label{eq:keeptopk}
TopK(v, k)_n = \begin{cases}
            v_n & \text{if $v_n$ is in the top $k$ of $v$.} \\
            -\infty & \text{otherwise.}
        \end{cases}.
\end{equation}
% --------Eq:keeptopk-----------
The individual experts $E_n$ in Eq. (\ref{eq:moe}) are neural networks. The gating network $G(x)$ incorporates sparse and noisy components before softmax, where $G(x)_n$ is the weight for expert $E_n$. In Eq. (\ref{eq:noise}), $H(x)$ introduces tunable Gaussian noise. $SN()$ denotes the standard normal distribution, $Softplus$ is the activation function, and $W_g$ and $W_{noise}$ are trainable weight matrices. $W_{noise}$ is the noise term for load balancing. $TopK(v, k)$ retains only the top $k$ values of $v$, setting the rest to $-\infty$ to ensure corresponding gate values become 0.

\subsubsection{Mixture of Noise Extractors (MoNE)}
The proposed Mixture of Noise Experts (MoNE) module adaptively captures diverse forgery traces by leveraging different types of noise extractors, as shown in Fig. \ref{fig:fat}(b). Specifically, we designate the previously mentioned noise extractors, namely HFConv \cite{li2019localization}, SRMConv \cite{fridrich2012rich}, BayarConv \cite{bayar2018constrained}, and CDConv \cite{yu2020searching}, as ${NE}_1$, ${NE}_2$, ${NE}_3$, and ${NE}_4$, respectively. Building upon these extractors, we construct multiple noise expert networks denoted as $\{{NE}_1, {NE}_2, {NE}_3, {NE}_4\}$. 
Furthermore, different noise experts can acquire overlapping knowledge or information, resulting in parameter redundancy and reduced focus within the expert networks \cite{dai2024deepseekmoe,rajbhandari2022deepspeed}. 
To address this issue, we introduce a Shared Expert ($SE$) in the form of a vanilla convolution layer. 
The purpose of $SE$ is to capture and consolidate the shared knowledge across different contexts, thereby alleviating parameter redundancy within the noise experts. This integration enhances the learning capacity of the noise experts for forged cues.
Consequently, the proposed MoNE module can be formulated as follows:
% --------Eq:mone-----------
\begin{equation}\label{eq:mone}
y = \sum_{n=1}^{4}G(x)_n{NE}_n(x) + SE(x).
\end{equation}
% --------Eq:mone-----------

Given that the original MoE layer is primarily designed for 1-D sequence data and not directly applicable to 2-D images, we propose an improvement to incorporate global information from the input $x$ within the $H(x)$ function. This is achieved by employing global average pooling denoted as $avg$ and linear layers represented as $F_g$ and $F_{noise}$. Consequently, the updated form of Eq. (\ref{eq:noise}) is as follows:
% --------Eq:noise2-----------
\begin{equation} \label{eq:noise2}
    \begin{aligned}
     H(x)_n = (x_g)_n + SN() \cdot Softplus((x_{noise})_n), \\
     s.t. \;\; x_g = F_g(avg(x)), x_{noise} = F_{noise}(avg(x)).
    \end{aligned}
\end{equation}
% --------Eq:noise2-----------
To process multi-face manipulation images using different noise extraction expert networks, we replace Eq. (\ref{eq:moe}) and (\ref{eq:noise}) with Eq. (\ref{eq:mone}) and (\ref{eq:noise2}). 
The gating network $G$ calculates weights by considering the global information of the input 2D feature map, resulting in $k$ gate values that dynamically assign the corresponding noise experts.
In this way, the proposed MoNE module adeptly selects and combines the outputs of diverse noise extractors, effectively harnessing their complementary capabilities to learn discriminative forgery traces.
In our implementation, we set $k=4$, allowing the gating network to adaptively control four noise experts for handling different samples. This enables the extraction of more generalized forgery patterns.

%--------------Figure Visualization---------------
\begin{figure*}[!t]
\centering
\includegraphics[width=0.99\textwidth]{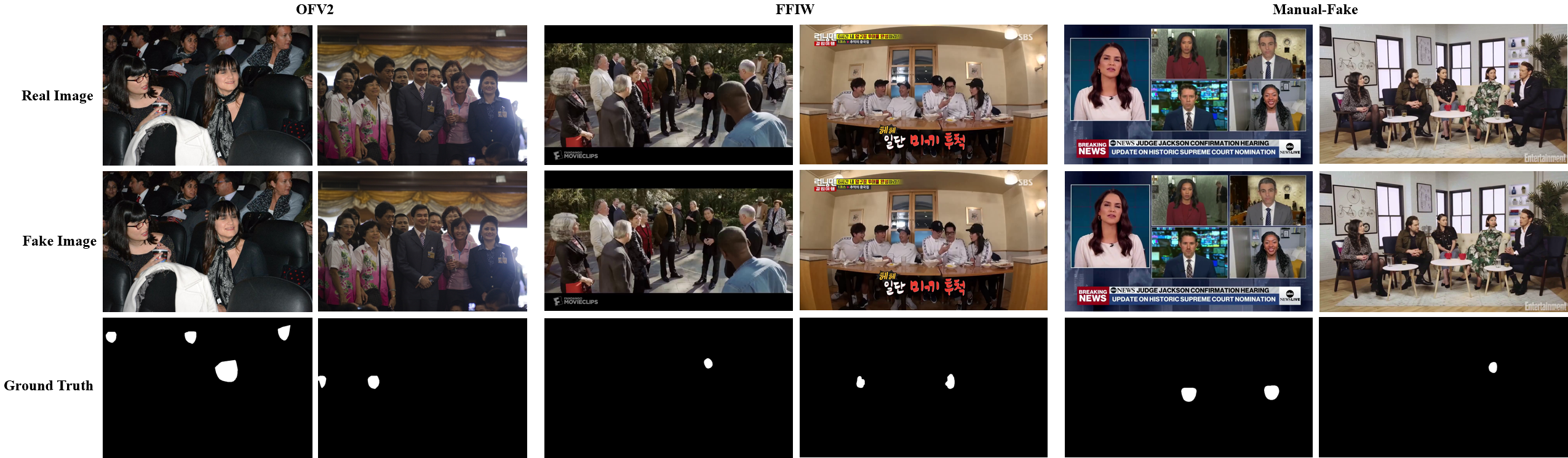}
\caption{We present the collected datasets, namely OFV2, FFIW, and Manual-Fake. Row `Real Image' represents genuine samples, row `Fake Image' represents forged samples (one or more faces tampered with), and row `Ground Truth' represents annotations of the tampered regions.
}
\label{fig:supp_datasets}
\end{figure*}
%--------------Figure Visualization---------------

\subsubsection{Importance Loss}
During training, the gating network often converges to a biased state, consistently assigning large weights to a small subset of experts \cite{eigen2013learning,bengio2015conditional,shazeer2017outrageously}. 
Following \cite{shazeer2017outrageously}, we employ an Importance Loss term $L_{im}$, calculated as the square of the coefficient of variation ($CV$) of the importance values, multiplied by a scaling factor $w_{im}$.
Mathematically, we have:
% --------Eq:importanceloss-----------
\begin{equation}\label{eq:imloss}
L_{im} = w_{im} \cdot CV(\sum_{x \in B}G(x))^2,
\end{equation}
% --------Eq:importanceloss-----------
where $B$ is the batch size of the training phase. 
This additional loss encourages a more balanced contribution from each expert in the network.
In the multi-level cues process, the overall MoNE loss is as follows:
% --------Eq:moneloss-----------
\begin{equation}\label{eq:moneloss}
L_{mone} = L_{im}^0 + L_{im}^1 + L_{im}^2 + L_{im}^3.
\end{equation}
% --------Eq:moneloss-----------

\subsection{Loss Functions}
\label{sec_loss}
We begin by applying a cross-entropy (CE) loss function, denoted as $L_{img}$, to classify authentic or manipulated faces.
To address the imbalance between genuine and manipulated pixel categories, we propose a sample-level weighted CE function for localization prediction in MoNFAP. The pixel-level loss ($L_{pix}$) is defined as follows:
% --------Eq:pix_loss1-----------
\begin{equation} \label{eq:pix_loss1}
   L_{pix} = CE_{genuine} + \lambda \cdot CE_{manipulated},
\end{equation}
% --------Eq:pix_loss1-----------
Here, $CE_{genuine}$ calculates the CE function for genuine samples only, while $CE_{manipulated}$, conversely, does so exclusively for manipulated samples.
The weighting factor $\lambda$ balances the contribution of the two categories at the sample level. In our implementation, we set $\lambda = 10$.
For the auxiliary localizer that generates masked attention, we define the loss in the same manner as $L_{aux}$.

For the MoNE importance loss ($L_{mone}$) is described in Sec. \ref{sec_MNP}.
Finally, the multi-task loss function $Loss$ is used to jointly optimize the model parameters, we have:
% --------Eq:loss-----------
\begin{equation} \label{eq:loss}
   Loss = L_{img} + L_{pix} + L_{aux} + L_{mone}.
\end{equation}
% --------Eq:loss-----------

\section{Benchmarks}
\subsection{Datasets}
We collect new multi-face data and select multi-face images from the existing datasets, as shown in Tab. \ref{tab:datasets} and Fig. \ref{fig:supp_datasets}.

% --------tab:datasets-----------
\begin{table}[!t]
\caption{Statistical of multi-face manipulation datasets, including fake and corresponding real images for each subset.}
\label{tab:datasets}
\centering
\begin{tabular}{ccccc}
\toprule[1pt]
Datasets    & Train & Validation   & Test  & Total  \\ 
\cmidrule(r){1-1} \cmidrule(r){2-2} \cmidrule(r){3-3} \cmidrule(r){4-4} \cmidrule(r){5-5}
OFV2        & $79,462$ & $13,984$ & $35,104$ & $128,550$ \\
FFIW        & $84,612$ & $4,308$  & $33,252$ & $122,172$ \\
Manual-Fake & --    & --    & $19,794$ & $19,794$  \\ 
\bottomrule[1pt]
\end{tabular}
\vspace{-0.3cm}
\end{table}
% --------tab:datasets-----------

\subsubsection{OFV2}
% The OpenForensics \cite{le2021openforensics} dataset is first proposed to support the advancement of multi-face forgery detection and segmentation tasks, which provides rich pixel-level forgery regions annotation.
% GAN model generates the manipulated face \cite{shen2020interpreting,pidhorskyi2020adversarial} and after a complex post-processing process the final data is obtained as $44,122$ training images, $7,308$ validation images, and $18,895$ test-development images.
% The OpenForensics \cite{le2021openforensics} contains various types of image scenes and a large number of high-resolution multi-face images.
% Despite its merits, the OpenForensics \cite{le2021openforensics} dataset suffers from a limitation: it solely comprises manipulated images and lacks corresponding genuine images, rendering it inadequate for image-level real-fake classification tasks.
% To address this gap, we gather the corresponding genuine images from the Open Images \cite{kuznetsova2020open} dataset while manually filtering out noisy and irrelevant samples to compose a new multi-face manipulation dataset OpenForensics-V2 (OFV2), as shown in Fig. \ref{fig:supp_datasets}.
The OpenForensics \cite{le2021openforensics} supports multi-face forgery segmentation, providing detailed pixel-level annotations for forgery regions. 
It includes $44,122$ training images, $7,308$ validation images, and $18,895$ test-development images generated by a GAN model \cite{shen2020interpreting,pidhorskyi2020adversarial} followed by complex post-processing. 
While it contains diverse high-resolution multi-face images, it has a limitation: it only includes manipulated images without corresponding genuine ones, making it unsuitable for image-level real-fake classification.
To address this, we collect genuine images from the Open Images \cite{kuznetsova2020open} dataset, manually filtering out noisy samples to create the new multi-face manipulation dataset, OpenForensics-V2 (OFV2), as shown in Fig. \ref{fig:supp_datasets}.

\subsubsection{FFIW}
% The FFIW \cite{zhou2021face} is a video-level face manipulation dataset that incorporates three deepfake methods \cite{faceswap2019,perov2020deepfacelab,nirkin2019fsgan} to generate diverse forged faces and provides pixel-level annotation, Within the FFIW dataset \cite{zhou2021face}, each forged video is accompanied by its corresponding genuine video. The dataset is split into distinct train ($16,000$ videos), validation ($500$ videos), and test ($3,500$ videos) sets
% However, it is worth noting that only some subsets of the FFIW dataset \cite{zhou2021face} comprise multi-face manipulations. 
% To address this limitation, we filter each video to identify the number of faces in each frame and sample up to 10 multi-face images (with at least two faces) at regular intervals. 
% This process results in creating a new multi-face manipulation image dataset derived from FFIW, as shown in Fig. \ref{fig:supp_datasets}.
The FFIW \cite{zhou2021face} includes video-level face manipulations using three deepfake methods \cite{faceswap2019,perov2020deepfacelab,nirkin2019fsgan}, providing pixel-level annotations. 
Each forged video is paired with its genuine counterpart, and the dataset is divided into training ($16,000$ videos), validation ($500$ videos), and test ($3,500$ videos) sets. 
However, only some video samples contain multi-face manipulations.
To address this, we filter videos to identify the number of faces per frame and sample up to 10 frames (with at least two faces) at regular intervals. This process creates a new multi-face manipulation image dataset derived from FFIW (see Fig. \ref{fig:supp_datasets}).

\subsubsection{Manual-Fake}
% The Manual-Fake \cite{haiwei2022exploring} is also a video-level face manipulation dataset that collects $1,000$ pristine videos and $1,000$ fake videos by five deepfake methods.
% Considering the pervasive influence of online social networks (OSNs) in propagating Deepfake videos, the Manual-Fake dataset \cite{haiwei2022exploring} encompasses transmitted versions of videos through the five most prominent OSN platforms (i.e., Facebook, WhatsApp, TikTok, YouTube, and WeChat).
% The Manual-Fake \cite{haiwei2022exploring} has the same problem as FFIW \cite{zhou2021face}, so we use the same processing and sampling methods to compose the new multi-face manipulation image dataset, as shown in Fig. \ref{fig:supp_datasets}.
% Considering that the Manual-Fake \cite{haiwei2022exploring} contains videos that are transmitted through OSNs, making it more representative of real-world scenarios, we utilize it as an unseen test set.
The Manual-Fake \cite{haiwei2022exploring} consists of 1,000 pristine and 1,000 fake videos generated by five deepfake methods. 
Given the influence of online social networks in spreading Deepfake videos, it includes versions transmitted through major platforms like Facebook, WhatsApp, TikTok, YouTube, and WeChat. 
Since it shares issues with FFIW, we apply the same processing and sampling methods to create a new multi-face manipulation image dataset (see Fig. \ref{fig:supp_datasets}). 
Due to its OSN-transmitted content, Manual-Fake serves as an unseen test set, enhancing its representation of real-world scenarios.

\begin{table*}[tb]
\begin{center}
\caption{Comparison with the SOTAs on the OFV2 and FFIW datasets. Types column represents different classification modes. 
}
\label{tab:intra-datasets}
\resizebox{0.95\linewidth}{!}
{
\begin{tabular}{ccccccccccc}
\toprule[1pt]
\multirow{2}{*}{Types} & \multirow{2}{*}{Methods} & \multirow{2}{*}{Reference} & \multicolumn{4}{c}{OFV2} & \multicolumn{4}{c}{FFIW} \\ 
\cmidrule(r){4-7} \cmidrule(r){8-11}
 &  &  & ACC & AUC & F1-f & IoU-f & ACC & AUC & F1-f & IoU-f \\ 
 \cmidrule(r){1-1} \cmidrule(r){2-2} \cmidrule(r){3-3} \cmidrule(r){4-5} \cmidrule(r){6-7} \cmidrule(r){8-9} \cmidrule(r){10-11}
\multirow{3}{*}{\begin{tabular}[c]{@{}c@{}}Detection by \\ Localization\end{tabular}} & ManTra-Net \cite{wu2019mantra} & CVPR 2019 & 77.60 & 98.48 & 83.69 & 71.96  & 59.12 & 78.73 & 72.70 & 57.11 \\
 & HPFCN \cite{li2019localization} & ICCV 2019 & 93.87 & 99.47 & 84.10 & 72.56 & 79.67 & 88.15 & 85.81 & 75.15 \\
 & MVSS \cite{chen2021image}& TPAMI 2021 & 94.80 & 98.71 & 82.44 & 70.12 & 84.33 & 92.65 & 88.24 & 78.95 \\ 
 \cmidrule(r){1-11}
\multirow{3}{*}{Two-branch} & CATNet \cite{kwon2022learning} & IJCV 2022 & 95.58 & 99.89 & 90.83 & 83.19 & 88.58 & \textbf{98.85} & 87.86 & 78.35 \\
 & DOAGAN \cite{islam2020doa} & CVPR 2020 & 89.17 & 97.97 & 84.51 & 73.17 & 82.91 & 94.37 & 87.44 & 77.68 \\
 & HiFiNet \cite{guo2023hierarchical} & CVPR 2023 & 94.01 & 99.76 & 85.62 & 74.85 & 91.32 & 98.56 & 85.22 & 74.24 \\ 
 \cmidrule(r){1-11}
\multirow{2}{*}{Unified} 
 & \textbf{MoNFAP-C} & -- & \textbf{99.10} & \textbf{99.91} & 94.82 & 90.15 & 92.15 & 98.03 & 90.80 & 83.15 \\ 
 & \textbf{MoNFAP-R} & -- & 95.54 & 99.71 & \textbf{94.85} & \textbf{90.20} & \textbf{92.86} & 98.27 & \textbf{91.62} & \textbf{84.54} \\ 
 % & \textbf{MoNFAP-H} & HRNet & 94.52 & 99.34 & 91.67 & 84.62 & \textbf{93.10} & \textbf{98.85} & \textbf{92.63} & \textbf{86.27} \\ 
 \bottomrule[1pt]
\end{tabular}
}
\end{center}
\end{table*}

\subsection{Baseline Models}
% We conduct a comprehensive benchmark for multi-face manipulation detection and localization, evaluating the performance of various state-of-the-art (SOTA) methods across diverse scenarios. 
% Our benchmark facilitates a rigorous and reproducible comparison by encompassing both quantitative and qualitative evaluations.
% To ensure a fair and unbiased assessment, we curate a broad selection of source code publicly available methods.
% These methods can be categorized into two types:
% 1) Detection by localization: HPFCN \cite{li2019localization}, ManTra-Net \cite{wu2019mantra}, and MVSS \cite{chen2021image}.
% 2) Dual-branch network: CATNet \cite{kwon2022learning}, DOA-GAN \cite{islam2020doa}, and HiFi-Net \cite{guo2023hierarchical}.
We conduct a comprehensive benchmark for multi-face manipulation detection and localization, evaluating various state-of-the-art (SOTA) methods across diverse scenarios. Our benchmark includes both quantitative and qualitative assessments for rigorous and reproducible comparisons. To ensure fairness, we curate a wide selection of publicly available source code methods, categorized into two types:
1) Detection by localization: HPFCN \cite{li2019localization}, ManTra-Net \cite{wu2019mantra}, and MVSS \cite{chen2021image}.
2) Dual-branch network: CATNet \cite{kwon2022learning}, DOA-GAN \cite{islam2020doa}, and HiFi-Net \cite{guo2023hierarchical}.

\subsection{Evaluation Protocols and Metrics}
\subsubsection{Evaluation Protocols}
To comprehensively evaluate the multi-face forgery detection and localization methods, we establish three evaluation protocols:
a) Intra-dataset: Models are trained and tested on the OFV2 and FFIW datasets.
% b) Cross-dataset: 
% Models are trained on OFV2 and FFIW and tested on the unseen Manual-Fake dataset. 
% It is worth noting that their data distribution is different.
% This evaluates the model's generalization ability to different face manipulation methods and data sources.
b) Cross-dataset: Models are trained on OFV2 and FFIW and tested on the unseen Manual-Fake dataset, assessing generalization to different face manipulation methods and data sources.
% c) Real-world perturbations: we introduce various perturbations to simulate real-world scenarios and natural contexts within the test sets of OFV2 and FFIW, following \cite{le2021openforensics}. 
% Specifically, we categorize the perturbations into five major components: color manipulation, edge manipulation, image corruption, convolution mask transformation, and external effects. These perturbation methods are then randomly combined and applied to enhance the test images.
c) Real-world perturbations: We introduce various perturbations to simulate real-world scenarios in the test sets of OFV2 and FFIW, categorized into five components: color, edge, image corruption, convolution mask transformation, and external effects. These perturbations are randomly combined and applied to enhance the test images, following \cite{le2021openforensics}.

\subsubsection{Evaluation Metrics}
For multi-face manipulation detection evaluation, we report Accuracy (ACC) and Area Under the Receiver Operating Characteristic Curve (AUC).
To assess the localization performance, 
we compute the F1-score (F1) and Intersection over Union (IoU) specifically for the fake class of the manipulated samples, denoted as F1-f and IoU-f, respectively.

\subsection{Implementation Details}
% We adopt ConvNeXtV2-atto \cite{woo2023convnext}, ResNet-50 \cite{he2016deep}, and HRNet \cite{wang2020hrnet} as backbone networks, namely MoNFAP-C, MoNFAP-R, and MoNFAP-H, respectively.
% for ablation study and comparison experiments, 
% We adopt ConvNeXtV2-atto \cite{woo2023convnext} and ResNet-50 \cite{he2016deep}as backbone networks, namely MoNFAP-C and MoNFAP-R, respectively.
% All MoNFAP models are trained using the AdamW optimizer, with an initial learning rate of $0.00006$, betas set to $(0.9, 0.999)$, and a weight decay of $0.01$. Training is facilitated by employing the "poly" learning rate policy, where the learning rate is decayed according to the formula $(1 - \frac{iter}{total_{iter}})^{0.9}$.
% Input images are resized to $512\times512$ dimensions. 
% Regarding other benchmark methods, we adhere to the training protocols outlined in their respective original papers, unless otherwise specified.
% % and a batch size of $64$ is utilized.
% Random horizontal flipping is the sole data augmentation technique during the training phase.
% Synchronized batch normalization implemented by Pytorch 2.0.1 is enabled during multi-GPU training.
We use ConvNeXtV2-atto \cite{woo2023convnext} and ResNet-50 \cite{he2016deep} as backbone networks, referred to as MoNFAP-C and MoNFAP-R, respectively. 
All MoNFAP models are trained with the AdamW optimizer, an initial learning rate of $0.00006$, betas of $(0.9, 0.999)$, and a weight decay of  $0.01$, utilizing a "poly" learning rate policy: \((1 - \frac{iter}{total_{iter}})^{0.9}\). Input images are resized to 512×512 pixels. For other benchmark methods, we follow the original training protocols unless specified otherwise. Random horizontal flipping is the only data augmentation used during training, and synchronized batch normalization from PyTorch 2.0.1 is enabled for multi-GPU training.

\begin{table*}[tb]
\begin{center}
\caption{The models train on the FFIW or OFV2 datasets and test on the unseen Manual-Fake dataset.
The italicized numbers indicate the average of the generalization results.
}
\label{tab:cross-datasets_1}
\resizebox{0.95\linewidth}{!}{
\begin{tabular}{ccccccccc|cccc}
\toprule[1pt]
\multirow{2}{*}{Methods} & \multicolumn{4}{c}{FFIW $\rightarrow$ Manual-Fake} & \multicolumn{4}{c|}{OFV2 $\rightarrow$ Manual-Fake} & \multicolumn{4}{c}{\textit{Average}} \\ 
\cmidrule(r){2-5} \cmidrule(r){6-9} \cmidrule(r){10-13}
& ACC & AUC & F1-f & IoU-f & ACC & AUC & F1-f & IoU-f & \textit{ACC} & \textit{AUC} & \textit{F1-f} & \textit{IoU-f} \\ 
\cmidrule(r){1-1} \cmidrule(r){2-3} \cmidrule(r){4-5} \cmidrule(r){6-7} \cmidrule(r){8-9} \cmidrule(r){10-11} \cmidrule(r){12-13}
ManTra-Net \cite{wu2019mantra} & 50.57 & 54.83 & 33.00 & 19.76 & 51.17 & 49.47 & 00.59 & 00.30 & 50.87 & 52.15 & 16.80 & 10.03 \\
HPFCN \cite{li2019localization} & 50.76 & 54.75 & 29.63 & 17.39 & 50.27 & 50.33 & 01.72 & 00.87 & 50.52 & 52.54 & 15.68 & 09.13 \\
% MVSS \cite{chen2021image} & 54.95 & 57.44 & 27.59 & 16.00 & 50.15 & 50.77 & 01.49 & 00.75 & \textit{52.55} & \textit{54.11} & \textit{14.54} & \textit{08.38} \\
MVSS \cite{chen2021image} & 54.95 & 57.44 & 27.59 & 16.00 & 50.76 & 49.72 & 02.83 & 01.43 & 52.86 & 53.58 & 15.21 & 08.72 \\
\cmidrule(r){1-13}
CATNet \cite{kwon2022learning} & 60.59 & 73.92 & 33.25 & 19.94 & 50.83 & 46.33 & 00.70 & 00.35 & 55.71 & 60.13 & 16.98 & 10.15 \\
% DOAGAN \cite{islam2020doa} & 52.28 & 55.86 & 26.29 & 15.13 & 50.54 & 44.80 & 00.23 & 00.12 & \textit{51.41} & \textit{50.33} & \textit{13.26} & \textit{07.63} \\
 DOAGAN \cite{islam2020doa} & 52.28 & 55.86 & 26.29 & 15.13 & 50.51 & 48.02 & 00.42 & 00.21 & 51.40 & 51.94 & 13.36 & 07.67 \\
HiFiNet \cite{guo2023hierarchical} & 60.37 & 73.37 & 27.50 & 15.94 & 38.59 & 34.94 & 01.12 & 00.56 & 49.48 & 54.16 & 14.31 & 08.25 \\ 
\cmidrule(r){1-13}
\textbf{MoNFAP-C} & 58.91 & 66.94 & 30.16 & 17.75 & 51.67 & 53.42 & 01.79 & 00.91 & 55.29 & 60.18 & 15.98 & 09.33 \\
\textbf{MoNFAP-R} & 60.70 & 67.88 & 33.91 & 20.41 & 53.42 & 55.27 & 03.17 & 01.61 & \textbf{57.06} & \textbf{61.58} & \textbf{18.54} & \textbf{11.01} \\
% \textbf{MoNFAP-H} & 60.27 & 69.33 & 41.43 & 26.13 & 49.87 & 51.20 & 04.42 & 02.26 & \textit{55.07} & \textit{60.27} & \textbf{\textit{22.93}} & \textbf{\textit{14.20}} \\
\bottomrule[1pt]
\end{tabular}
}
\end{center}
\end{table*}

\begin{table*}[tb]
\begin{center}
\caption{The models are trained on the FFIW and OFV2 datasets and tested on their test set with added unknown perturbations.
The italicized numbers indicate the average of the robustness results.
}
\label{tab:robust-datasets_2}
\resizebox{0.95\linewidth}{!}{
\begin{tabular}{ccccccccc|cccc}
\toprule[1pt]
\multirow{2}{*}{Methods} & \multicolumn{4}{c}{FFIW} & \multicolumn{4}{c|}{OFV2} & \multicolumn{4}{c}{\textit{Average}} \\ 
\cmidrule(r){2-5} \cmidrule(r){6-9} \cmidrule(r){10-13}
& ACC & AUC & F1-f & IoU-f & ACC & AUC & F1-f & IoU-f & \textit{ACC} & \textit{AUC} & \textit{F1-f} & \textit{IoU-f} \\ 
\cmidrule(r){1-1} \cmidrule(r){2-3} \cmidrule(r){4-5} \cmidrule(r){6-7} \cmidrule(r){8-9} \cmidrule(r){10-11} \cmidrule(r){12-13} 
ManTra-Net \cite{wu2019mantra} & 57.40 & 70.70 & 58.87 & 41.71 & 74.63 & 91.29 & 62.97 & 45.95 & 66.02 & 81.00 & 60.92 & 43.83 \\
HPFCN \cite{li2019localization} & 64.54 & 68.73 & 56.96 & 39.82 & 77.03 & 84.83 & 52.94 & 36.00 & 70.79 & 76.78 & 54.95 & 37.91 \\
% MVSS \cite{chen2021image} & 73.20 & 79.59 & 72.63 & 57.02 & 87.84 & 95.17 & 72.96 & 57.43 & \textit{80.52} & \textit{87.38} & \textit{72.80} & \textit{57.23} \\
MVSS \cite{chen2021image} & 73.20 & 79.59 & 72.63 & 57.02 & 77.45 & 84.61 & 38.61 & 23.93 & 75.33 & 82.10 & 55.62 & 40.48 \\
\cmidrule(r){1-13}
CATNet \cite{kwon2022learning} & 69.53 & 78.43 & 63.22 & 46.22 & 68.84 & 86.12 & 62.38 & 45.33 & 69.19 & 82.28 & 62.80 & 45.78 \\
% DOAGAN \cite{islam2020doa} & 72.78 & 83.87 & 67.69 & 51.17 & 78.91 & 94.40 & 60.05 & 42.91 & \textit{75.85} & \textit{89.14} & \textit{63.87} & \textit{47.04} \\
DOAGAN \cite{islam2020doa} & 72.78 & 83.87 & 67.69 & 51.17 & 73.51 & 92.17 & 64.33 & 47.42 & 73.15 & \textbf{88.02} & 66.01 & 49.30 \\
HiFiNet \cite{guo2023hierarchical} & 70.18 & 76.36 & 65.43 & 48.62 & 56.49 & 79.89 & 61.51 & 44.42 & 63.34 & 78.10 & 63.47 & 46.52 \\ 
\cmidrule(r){1-13}
\textbf{MoNFAP-C} & 72.77 & 82.59 & 78.20 & 64.20 & 81.53 & 91.16 & 80.22 & 66.97 & \textbf{77.15} & 86.88 & \textbf{79.21} & \textbf{65.59} \\
\textbf{MoNFAP-R} & 72.85 & 81.60 & 77.49 & 63.25 & 77.10 & 88.43 & 75.01 & 60.01 & 74.98 & 85.02 & 76.25 & 61.63 \\
% \textbf{MoNFAP-H} & 75.17 & 85.19 & 77.33 & 63.04 & 83.53 & 92.27 & 80.59 & 67.49 & \textit{79.35} & \textit{88.73} & \textit{78.96} & \textit{65.27} \\
\bottomrule[1pt]
\end{tabular}
}
\end{center}
\end{table*}

\section{Comparison Experiments}

\subsection{Intra-datasets Evaluation}
We start by evaluating benchmark methods' detection and localization performance on the FFIW and OFV2 datasets, representing a significant challenge aligned with real-world scenarios not extensively explored in prior literature.
In Table \ref{tab:intra-datasets}, HPFCN \cite{li2019localization} and ManTra-Net \cite{wu2019mantra} exhibit poor performance in image-level classification, particularly in terms of AUC metric on the FFIW dataset. This is attributed to the image-level results being a byproduct of the localization task, lacking corresponding design optimizations. 
Conversely, MVSS \cite{chen2021image} incorporates additional image-level loss supervision, enhancing image-level classification performance. 
Two-branch methods \cite{kwon2022learning,islam2020doa,guo2023hierarchical} improve image classification performance but show limited gains in localization. 
For instance, on the FFIW dataset, CATNet \cite{kwon2022learning} achieves a classification AUC of $98.85\%$ with a larger backbone, yet its localization performance (IoU-f) is $4.8\%$ lower than our lightweight MoNFAP-C.
This discrepancy arises from its lack of leveraging the interactive information between the two tasks.
Our MoNFAP framework utilizes a token learning strategy to simultaneously produce classification and localization results, fully integrating classification information into the localizer and improving the localization performance. 

% Meanwhile, our method exhibits varying performance across different backbone networks. For instance, the lightweight ConvNeXt \cite{woo2023convnext} is more adept at fitting the OFV2 dataset, while the small object data in the FFIW dataset necessitates the high resolution provided by HRNet \cite{wang2020hrnet}. 
% Regardless of the backbone network employed, our approach significantly outperforms existing state-of-the-art methods regarding localization capabilities through the proposed multi-scale FAT and noise cues.

\subsection{Generalization to Cross-datasets}
We assess the generalization capability of models through cross-dataset experiments, i.e., training on the FFIW or OFV2 datasets and testing on unseen Manual-Fake dataset. 
The unseen datasets mean that used anonymous forgery methodologies based on unknown source data and it presents a challenging scenario for evaluating model performance.
Table \ref{tab:cross-datasets_1} presents the results of these cross-dataset experiments, offering valuable insights into the models' ability to generalize to unseen data distributions and forgery techniques. 
All methods display a significant performance decrease. 
The generalization ability of the model varies depending on the training set. For instance, when the model is trained on the OFV2 dataset, it performs poorly when tested on the unseen Manual-Fake dataset. This indicates a significant distribution discrepancy between the OFV2 and Manual-Fake datasets. Specifically, the Manual-Fake dataset predominantly consists of news broadcasting scenarios, whereas the OFV2 dataset contains almost no data of this nature.
Therefore, unseen data remains a substantial challenge for multi-face manipulation localization methods.
Our MoNFAP outperforms other state-of-the-art methods, particularly in terms of averaged localization performance, attributed to the specially designed FUP and MNM modules.

%--------------Figure Visualization---------------
\begin{figure*}[!t]
\centering
\includegraphics[width=0.99\textwidth]{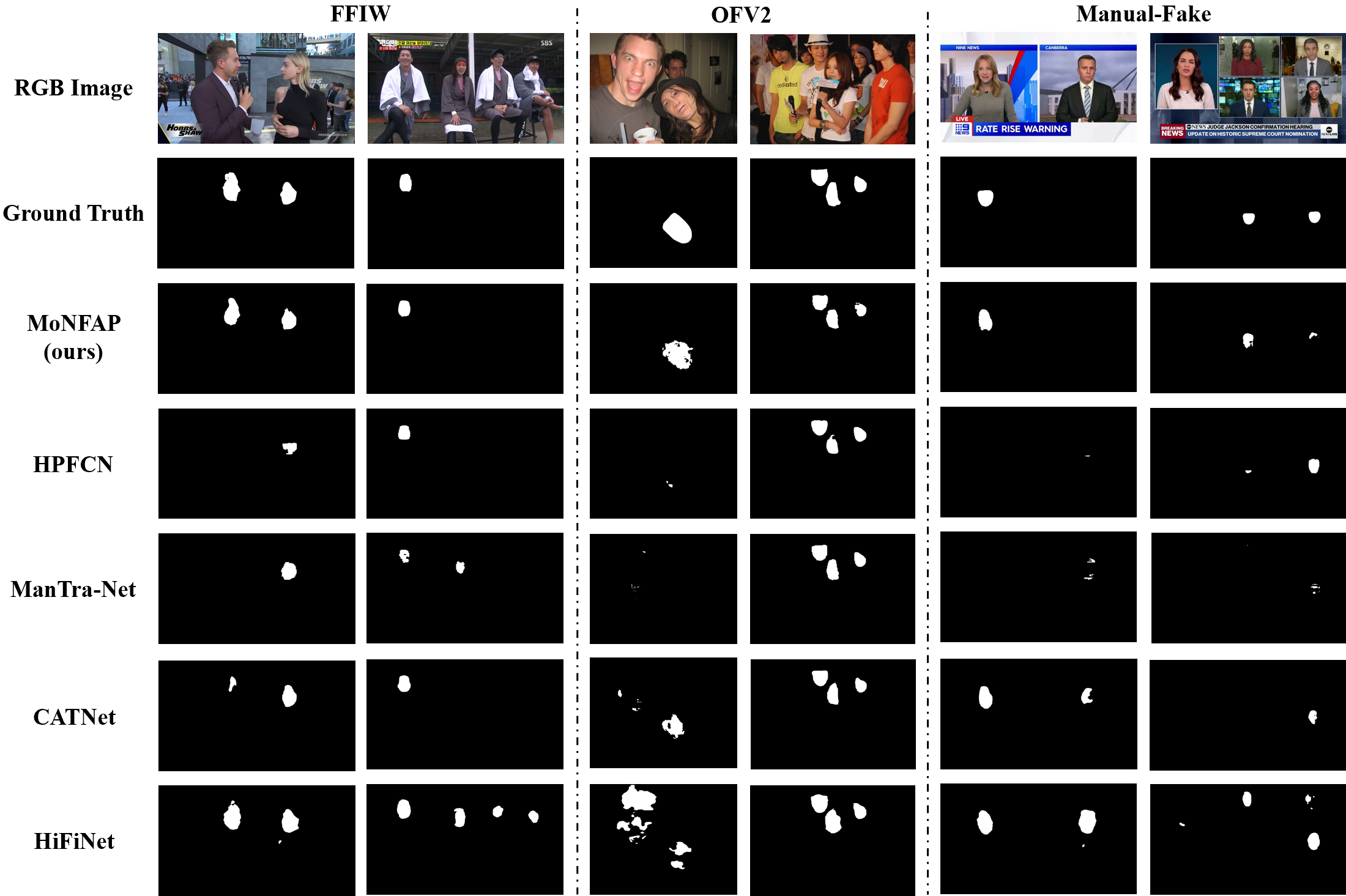}
\caption{Visualization of localization results on the FFIW, OFV2, and Manual-Fake datasets. 
Samples are randomly sampled from the test sets of FFIW, OFV2, and Manual-Fake, with the models trained on the respective training sets of FFIW and OFV2. 
The `RGB Image' row represents the input samples, the `Ground Truth' row represents the labels of the tampered regions. The remaining rows represent the prediction results of different models.
}
\label{fig:supp_vis_1}
\end{figure*}
%--------------Figure Visualization---------------

\subsection{Robustness to Real-world Perturbations}
The presence of manipulated images in real-world settings introduces various perturbations, disrupting manipulation traces and increasing the difficulty of detection and localization. 
To evaluate model robustness, we introduced a range of noise and blur operations to the test set of OFV2 and FFIW datasets, simulating real-world environments.
Among the methods evaluated in Table \ref{tab:robust-datasets_2}, many methods exhibit the most significant localization performance degradation on unseen perturbed data. 
In AUC performance for detection, our method outperforms models in the Detection by Localization categories \cite{li2019localization,wu2019mantra,chen2021image} and achieves results comparable to those in the Two-Branch categories \cite{kwon2022learning,islam2020doa,guo2023hierarchical}. However, our MoNFAP significantly surpasses existing state-of-the-art methods in localization performance.
This improvement is attributed to our approach innovatively incorporates the MoE concept to learn multi-scale mixed noise cues, thereby enhancing the model's robustness in localization tasks.
% .......
% This could be attributed to its failure to extract robust noise information from multi-level features. 
% In contrast, our approach innovatively incorporates the MoE concept to learn multi-scale noise cues, enhancing the FUP module's robustness in terms of localization.

\subsection{Visualization Experiment}
\label{sec:vis}
As shown in Fig. \ref{fig:supp_vis_1}, we visualize the localization prediction masks on the FFIW and OFV2 datasets. The samples in the FFIW column are randomly sampled from the test set of the FFIW dataset, with the model trained on the FFIW training set. The same applies to the OFV2 column. The visualization results of different methods indicate that our MoNFAP method is capable of better identifying multiple forged faces and minor tampered regions, while other methods exhibit poorer performance. For instance, HiFiNet \cite{guo2023hierarchical} not only predicts the forged facial regions, but also erroneously locates genuine facial features, indicating its overfitting to facial characteristics without learning the forgery clues.

To demonstrate the model's generalization ability, we visualize the localization prediction results on the unseen Manual-Fake dataset, as shown in Fig. \ref{fig:supp_vis_1}. 
It is worth noting that the model is trained on the FFIW dataset and tested solely on the unseen Manual-Fake dataset. 
The visualization results indicate that our method exhibits certain localization capabilities on unseen data, but further improvements are needed for small target forgery regions.

%--------------Table Extend Experiment---------------
\begin{table}
\begin{center}
\caption{Extend experiment on image forgery datasets. The model is trained on the CASIAv2 \cite{casiav2} dataset while tested on the other five datasets. The evaluation metric is the pixel-level F1 score.
}
\label{tab:image_forgery}
\resizebox{0.99\linewidth}{!}{
\begin{tabular}{cccccc|c}
\toprule[1pt]
Methods & COVER & Columbia & NIST16 & CASIAv1 & IMD2020 & \textit{Average} \\ \midrule
Mantra-Net \cite{wu2019mantra} & 09.0 & 24.3 & 10.4 & 12.5 & 05.5 & 12.3 \\
MVSS-Net \cite{chen2021image} & 25.9 & 38.6 & 24.6 & 53.4 & 27.9 & 34.1 \\
ObjectFormer \cite{wang2022objectformer} & 29.4 & 33.6 & 17.3 & 42.9 & 17.3 & 28.1 \\
PSCC-Net \cite{liu2022pscc} & 23.1 & 60.4 & 21.4 & 37.8 & 23.5 & 33.3 \\
NCL-IML \cite{zhou2023pre} & 22.5 & 44.6 & 26.0 & 50.2 & 23.7 & 33.4 \\ \midrule
\textbf{MoNFAP-R} & 26.34 & 44.66 & 26.92 & 59.12 & 27.14 & \textbf{36.84} \\
\bottomrule[1pt]
\end{tabular}
}
\end{center}
\end{table}
%--------------Table Extend Experiment---------------

\subsection{Extend Experiment} \label{sec:experiment_2_1}
To further validate the effectiveness of our approach, we conducted experiments on the image forgery datasets following \cite{ma2024imdlbenco}.
In general, the models are trained on the CASIAv2 \cite{casiav2} dataset, and tested on five unseen testing datasets including CASIAv1 \cite{casiav1}, COVER \cite{2016COVERAGE}, IMD2020 \cite{novozamsky2020imd2020}, NIST16 \cite{guan2019mfc}, and Columbia \cite{hsu2006detecting}.
As shown in  Table \ref{tab:image_forgery}, the localization results for other methods are sourced from \cite{ma2024imdlbenco}, with evaluation metrics focused solely on the pixel-level F1 score of manipulated images. 
The results indicate that our MoNFAP significantly outperforms other traditional image tampering localization methods across all five unseen datasets, demonstrating the applicability of our model to conventional manual image editing techniques.

\section{Ablation Studies}
To save resources and accelerate training speed, we choose the lightweight ConvNeXtV2-atto \cite{woo2023convnext} as the backbone network, the models train and test on FFIW \cite{zhou2021face} datasets, with image-level ACC and AUC and pixel-level F1-f and IoU-f as evaluation metrics.

\subsection{Analysis on the MoNFAP}
\label{sec_Fab_1}
\subsubsection{Impact of the Proposed Modules}
Table \ref{tab:ab_modules} presents our experiments analyzing different modules in MoNFAP.
`baseline' refers to the FCN method using ConvNeXtV2-atto \cite{woo2023convnext} as the backbone network. `+FUP (w/o MSS)' represents the baseline model with only the FUP module and no multi-scale strategy. `+FUP' denotes the baseline model with both the FUP module and the multi-scale strategy. `+FUP+MNM' signifies the final proposed MoNFAP framework.
The FUP module and multi-scale strategy improve performance compared to the baseline model, particularly in terms of localization capability. 
Additionally, the MoNP enhances both classification and localization performance, demonstrating the effectiveness of mixture noise cues.

\subsubsection{Task Mode}
Table \ref{tab:ab_modes} illustrates four different classification task modes, constructed to provide a fair evaluation of their characteristics.
`global statistics' refers to the classification result obtained by selecting the maximum value from the FUP-predicted localization mask, without classification loss. 
`additional loss' indicates that during the training process, image-level classification loss supervision based on the maximum value of the FUP-predicted localization mask is added, and it is the same as `global statistics' during testing.
The two modes above are collectively referred to as `detection by localization', as shown in Fig. \ref{fig:motivation_1}(a).
`two-branch' signifies the application of an independent classification branch outside of FUP, and the output tokens in FUP are not considered in the classification result.
Based on the results, the classification performance of `additional loss' is better than `global statistics', due to the effect of classification supervision. 
Although `two-branch' achieved excellent classification results, the lack of interaction between the two tasks resulted in no improvement in localization performance.
Our `token learning' shows a similar classification performance to `two-branch', but our localization performance exceeds it by $1.51\%$ in terms of IoU-f.
This indicates our method's effective enhancement of the localizer's capability.

\subsubsection{Number of Feature Scales}
\begin{figure}[!t]
\centering
\includegraphics[width=0.45\textwidth]{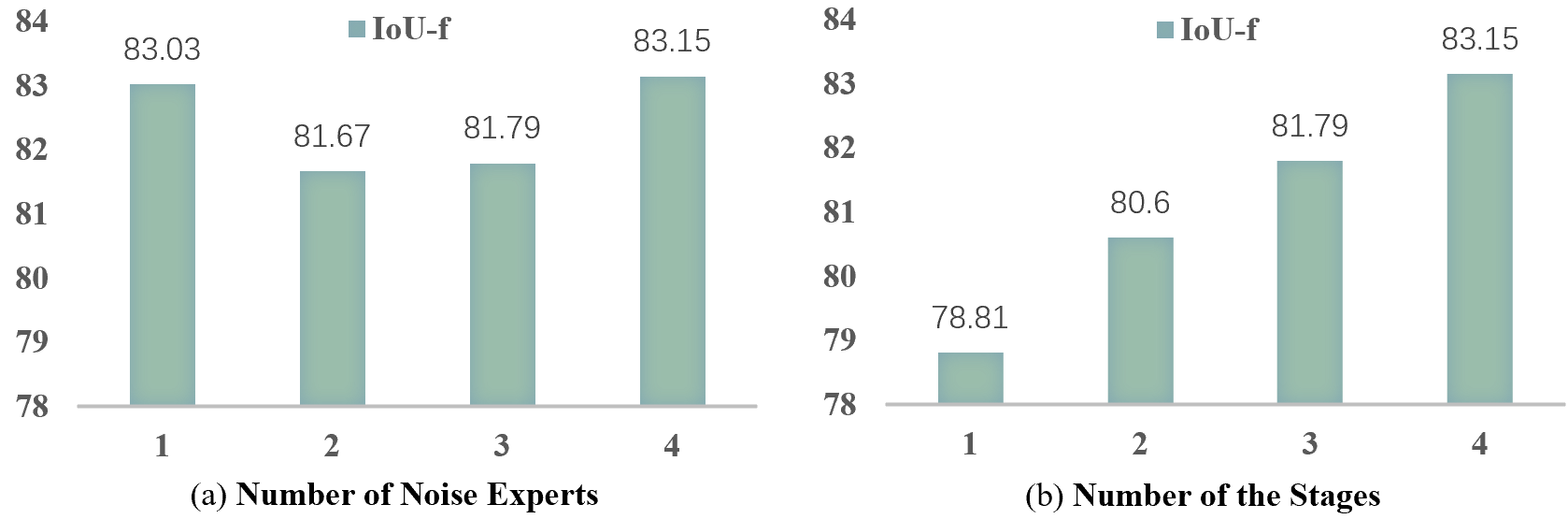}
\caption{We construct experimental analysis on different the number of noise experts in the MoNE and the number of stages in the MSS.
}
\label{fig:ab_1}
\end{figure}

As shown in Fig. \ref{fig:ab_1}(b), we conduct experiments with different numbers of feature scales. 
From the experimental results, it is evident that the performance is optimal when using 4 scales, indicating that multi-scale features contribute to improving the model's localization ability.

\begin{table}[tb]
\begin{center}
\caption{Analysis on the proposed the proposed modules} 
\label{tab:ab_modules}
\resizebox{0.90\linewidth}{!}{
        \begin{tabular}{ccccc}
            \toprule[1pt]
            Models & ACC & AUC & F1-f & IoU-f \\
            \cmidrule(r){1-1} \cmidrule(r){2-3} \cmidrule(r){4-5} 
            baseline & 91.68 & 97.80 & 86.73 & 76.57 \\
            +FUP (w/o MSS) & 91.88 & 98.05 & 89.40 & 80.83 \\
            +FUP & 91.82 & 97.78 & 90.29 & 82.30 \\
            \textbf{+FUP+MNM} & 92.15 & 98.03 & 90.80 & 83.15  \\
            \bottomrule[1pt]
        \end{tabular}
        }
\end{center}
\end{table}

\begin{table}[tb]
\begin{center}
\caption{Different classification task modes} 
\label{tab:ab_modes}
\resizebox{0.90\linewidth}{!}{
        \begin{tabular}{ccccc}
            \toprule[1pt]
            Types & ACC & AUC & F1-f & IoU-f \\
            \cmidrule(r){1-1} \cmidrule(r){2-3} \cmidrule(r){4-5}
            global statistics & 80.85 & 79.50 & 89.33 & 80.71 \\
            additional loss & 88.60 & 95.10 & 89.93 & 81.71 \\
            two-branch & 92.13 & 98.12 & 89.89 & 81.64  \\
            \textbf{tokens learning} & 92.15 & 98.03 & 90.80 & 83.15  \\
             \bottomrule[1pt]
        \end{tabular}
        }
\end{center}
\end{table}

%--------------Figure ablation studies---------------
%--------------Figure ablation studies---------------

\subsection{Analysis on the MoNE}
\subsubsection{Number of Noise Experts}
We compared different variants to find the optimal number of the noise extractors,  as shown in Fig. \ref{fig:ab_1}(a).
We observed that the variant with 4 different noise extractors outperformed the others in terms of localization performance.
Thanks to the weight allocation advantage of the Gating network, the 4 different noise extractors handle different samples in a batch in an optimal way to model forgery clues.
Other quantities of variants learn features that are not comprehensive and general enough, leading to poorer performance.

\begin{table}[!t]
\begin{center}
\caption{The number of shared experts (SE)} 
\label{tab:ab_SE}
\resizebox{0.70\linewidth}{!}{
        \begin{tabular}{ccccc}
            \toprule[1pt]
            SE & ACC & AUC & F1-f & IoU-f \\
            \cmidrule(r){1-1} \cmidrule(r){2-3} \cmidrule(r){4-5}
            0 & 91.92 & 97.93 & 90.04 & 81.89 \\ 
            \textbf{1} & 92.15 & 98.03 & 90.80 & 83.15 \\
            2 & 91.93 & 98.19 & 90.80 & 81.98 \\
            3 & 91.17 & 97.99 & 89.92 & 81.69 \\ 
            \bottomrule[1pt]
        \end{tabular}
        }
\end{center}
\end{table}

\begin{table}[!t]
\begin{center}
\caption{Methods related to MoNE} 
\label{tab:ab_MoNE}
\resizebox{0.80\linewidth}{!}{
        \begin{tabular}{ccccc}
            \toprule[1pt]
            Types & ACC & AUC & F1-f & IoU-f \\
            \cmidrule(r){1-1} \cmidrule(r){2-3} \cmidrule(r){4-5}
            Add & 91.90 & 97.91 & 90.40 & 82.47 \\
            Cat & 91.29 & 97.67 & 89.71 & 81.34 \\
            MoE & 90.41 & 98.03 & 90.52 & 82.68 \\
            \textbf{MoNE} & 92.15 & 98.03 & 90.80 & 83.15 \\
             \bottomrule[1pt]
        \end{tabular}
        }
\end{center}
\end{table}

\subsubsection{Number of Shared Experts}
Shared experts can learn redundant knowledge and alleviate the learning burden of different noise experts. As shown in Tab. \ref{tab:ab_SE}, one shared expert is optimal. 
A value of '0' indicates no shared experts, resulting in knowledge redundancy among the noise experts and poor performance. 
On the other hand, an excessive number of experts leads to increased parameters and optimization difficulties.

\subsubsection{Analysis of Structure Similar to MoNE}
As shown in Tab. \ref{tab:ab_MoNE}, to demonstrate the advantages of MoNE, we conducted ablation experiments with a similar structure. 'Add' indicates the element-wise addition of four different noises, 'Cat' indicates the concatenation of four different noises along the channel dimension, and 'MoE' represents the original mixed expert structure. Our MoNE allocates adaptive weights to different noise experts to handle different samples, integrating their respective advantages and outperforming other methods.

\subsection{Analysis on Importance Loss Function}
As shown in Table \ref{tab:ab_moneloss}, `w/o $L_{mone}$ loss' denotes the absence of the Importance Loss, while `w/ $L_{mone}$ loss' represents the opposite. The results indicate that the $L_{mone}$ loss enhances the localization performance, attributed to its ability to balance the competition among multiple noise experts and stabilize the training process.

\subsection{Analysis on Weighting Factor $\lambda$}
In Eq. (\ref{eq:loss}), the weighting factor $\lambda$ is used to adjust the different weights of the localization loss for real and fake samples, where $\lambda = 1$ indicates equal weights for the localization loss of real and fake samples, and $\lambda$ greater than $1$ indicates a larger weight for the localization loss of fake samples. 
As shown in Table \ref{tab:ab_factor}, the optimal performance is achieved when $\lambda = 10$, therefore, we choose it as the parameter for other experiments.

\begin{table}[!t]
\begin{center}
\caption{Ablation experiments of the importance loss function.} 
\label{tab:ab_moneloss}
 \resizebox{0.95\linewidth}{!}{
        \begin{tabular}{ccccc}
            \toprule[1pt]
            Types & ACC & AUC & F1-f & IoU-f \\
            \cmidrule(r){1-1} \cmidrule(r){2-3} \cmidrule(r){4-5}
            w/o $L_{mone}$ loss & 92.05 & 98.06 & 90.47 & 82.59 \\
            \textbf{w/ $L_{mone}$} loss & 92.15 & 98.03 & 90.80 & 83.15  \\
             \bottomrule[1pt]
        \end{tabular}
        }
\end{center}
\end{table}

\begin{table}[!t]
\begin{center}
\caption{The number of weighting factor $\lambda$.} 
\label{tab:ab_factor}
 \resizebox{0.75\linewidth}{!}{
        \begin{tabular}{ccccc}
            \toprule[1pt]
            $\lambda$ & ACC & AUC & F1-f & IoU-f \\
            \cmidrule(r){1-1} \cmidrule(r){2-3} \cmidrule(r){4-5}
            1 & 92.57 & 98.49 & 88.41 & 79.22 \\ 
            5 & 90.38 & 97.63 & 89.39 & 80.81 \\ 
            \textbf{10} & 92.15 & 98.03 & 90.80 & 83.15 \\
            15 & 91.36 & 97.44 & 90.43 & 82.53 \\
            20 & 9101 & 97.11 & 90.23 & 82.20 \\ 
            \bottomrule[1pt]
        \end{tabular}
        }
\end{center}
\end{table}

\begin{table}[!t]
\begin{center}
\caption{The number of the binary threshold of the masked attention map.} 
\label{tab:ab_th}
 \resizebox{0.80\linewidth}{!}{
        \begin{tabular}{ccccc}
            \toprule[1pt]
            Threshold& ACC & AUC & F1-f & IoU-f \\
            \cmidrule(r){1-1} \cmidrule(r){2-3} \cmidrule(r){4-5}
            0.0 & 91.25 & 97.40 & 89.88 & 81.62 \\
            0.3 & 91.87 & 97.83 & 90.35 & 82.40 \\
            \textbf{0.5} & 92.15 & 98.03 & 90.80 & 83.15 \\
            0.7 & 91.89 & 97.94 & 90.06 & 81.92 \\
            0.9 & 91.75 & 97.87 & 89.97 & 81.77 \\
             \bottomrule[1pt]
        \end{tabular}
        }
\end{center}
\end{table}

\subsection{Analysis on the Threshold of the Masked Attention Map}
The masked cross-attention method in the FAT module utilizes an additional localization layer to provide the masked attention map, as shown in Eq. (\ref{eq:masked-attn}). 
We conduct experiments with different thresholds for the binarization of the mask, as shown in Table \ref{tab:ab_th}. 
Here, $0$ indicates the absence of the masked-attention strategy, while $0.3$, $0.5$, $0.7$, and $0.9$ represent different binarization thresholds. 
The experimental results indicate that a threshold of $0.5$ yields the optimal performance.

\section{Conclusion and Discussion}
\subsection{Conclusion}
This paper introduces \textit{MoNFAP}, a Mixture-of-Noises Enhanced Forgery-Aware Unified Predictor, addressing the gap in previous research on multi-face forgery detection and localization within the broader forgery research community.
We propose a token learning strategy and a Forgery-Aware Transformer module to jointly predict the classification and positioning results by reasoning the relationship between real-fake tokens and image features. 
This process effectively enhances the localizer's capability by incorporating classification information.
Furthermore, we introduce a Mixture-of-Noises Module that utilizes the concept of a mixture of experts. This module aggregates different types of noise cues, enhancing generalized RGB features.
Finally, we establish a comprehensive benchmark to evaluate state-of-the-art methods, and the proposed \textit{MoNFAP} achieves significant performance.

\subsection{Discussion}
Currently, there are no methods for simultaneous pixel-level localization and image-level detection in the multi-face forgery research community. This paper introduces benchmarks for both tasks and proposes a novel joint prediction method. We aim to advance the field of multi-face forgery localization.

% However, our work does not include the latest diffusion-generated data. In future work, we aim to construct a large-scale multi-face forgery dataset based on diffusion techniques to enhance research in this field.

% \subsection{Social Impact}
% The multi-face dataset utilized in this study is derived from Open Images \cite{kuznetsova2020open}, OpenForensics \cite{le2021openforensics}, FFIW \cite{zhou2021face}, and Manual-Fake \cite{haiwei2022exploring}. 
% It is worth noting that all the images we used strictly comply with the licenses and regulations established by each source dataset.
The multi-face dataset in this study is sourced from Open Images\cite{kuznetsova2020open}, OpenForensics \cite{le2021openforensics}, FFIW \cite{zhou2021face}, and Manual-Fake \cite{haiwei2022exploring}. All images comply with the licenses and regulations of their respective datasets.

\bibliographystyle{IEEEtran}
\bibliography{bib/dfd,bib/iml,bib/mfl,bib/others,bib/sfl}

\end{document}